\PassOptionsToPackage{table}{xcolor}
\documentclass[manuscript,screen,preprint,nonacm]{acmart}

\AtBeginDocument{%
  \providecommand\BibTeX{{%
    \normalfont B\kern-0.5em{\scshape i\kern-0.25em b}\kern-0.8em\TeX}}}

\newcommand{\tick}[0]{\multicolumn{1}{l}{$\checkmark$}}
\newcommand{\namedpar}[1]{\noindent \textbf{#1}}

\usepackage{longtable}
\usepackage{tabularx, booktabs}
\usepackage{tcolorbox}

\setcopyright{acmlicensed}

\copyrightyear{2024}
\acmYear{2024}
\setcopyright{rightsretained}
\acmConference[FAccT '24]{The 2024 ACM Conference on Fairness, Accountability, and Transparency}{June 3--6, 2024}{Rio de Janeiro, Brazil}
\acmBooktitle{The 2024 ACM Conference on Fairness, Accountability, and Transparency (FAccT '24), June 3--6, 2024, Rio de Janeiro, Brazil}\acmDOI{10.1145/3630106.3658931}
\acmISBN{979-8-4007-0450-5/24/06}

\begin{document}

\title{Lazy Data Practices Harm Fairness Research}

\author{Jan Simson}
\affiliation{%
  \institution{LMU Munich}
  \city{Munich}
  \country{Germany}}
\affiliation{%
  \institution{Munich Center for Machine Learning (MCML)}
  \city{Munich}
  \country{Germany}}
\email{jan.simson@lmu.de}

\author{Alessandro Fabris}
\affiliation{%
  \institution{Max Planck Institute for Security and Privacy}
  \city{Bochum}
  \country{Germany}}
\email{alessandro.fabris@mpi-sp.org}

\author{Christoph Kern}
\affiliation{%
  \institution{LMU Munich}
  \city{Munich}
  \country{Germany}}
\affiliation{%
  \institution{Munich Center for Machine Learning (MCML)}
  \city{Munich}
  \country{Germany}}
\affiliation{%
  \institution{University of Maryland}
  \city{College Park}
  \country{USA}}
\email{christoph.kern@lmu.de}


\begin{abstract}
  Data practices shape research and practice on fairness in machine learning (fair ML). Critical data studies offer important reflections and critiques for the responsible advancement of the field by highlighting shortcomings and proposing recommendations for improvement. In this work, we present a comprehensive analysis of fair ML datasets, demonstrating how unreflective yet common practices hinder the reach and reliability of algorithmic fairness findings. We systematically study protected information encoded in tabular datasets and their usage in 280 experiments across 142 publications.
  
  Our analyses identify three main areas of concern: (1) a \textbf{lack of representation for certain protected attributes} in both data and evaluations; (2) the widespread \textbf{exclusion of minorities} during data preprocessing; and (3) \textbf{opaque data processing} threatening the generalization of fairness research. By conducting exemplary analyses on the utilization of prominent datasets, we demonstrate how unreflective data decisions disproportionately affect minority groups, fairness metrics, and resultant model comparisons.  Additionally, we identify supplementary factors such as limitations in publicly available data, privacy considerations, and a general lack of awareness, which exacerbate these challenges. To address these issues, we propose a set of recommendations for data usage in fairness research centered on transparency and responsible inclusion. This study underscores the need for a critical reevaluation of data practices in fair ML and offers directions to improve both the sourcing and usage of datasets.
\end{abstract}

\ccsdesc[500]{Social and professional topics~User characteristics}
\ccsdesc[500]{Computing methodologies~Machine learning}

\keywords{critical data studies, protected groups, fair ML generalization, reproducibility}


\maketitle

\begin{tcolorbox}[colframe=red, colback=white, arc=2mm, boxrule=1mm, width=\textwidth]

\textbf{When referencing this work, please cite it as follows:}

Jan Simson, Alessandro Fabris, and Christoph Kern. 2024. Lazy Data Practices Harm Fairness Research. In \textit{The 2024 ACM Conference on Fairness, Accountability, and Transparency (FAccT ’24), June 3–6, 2024, Rio de Janeiro, Brazil}. ACM, New York, NY, USA, 25 pages. \href{https://doi.org/10.1145/3630106.3658931}{https://doi.org/10.1145/3630106.3658931}

\end{tcolorbox}

\section{Introduction}

The identification and mitigation of harms against vulnerable individuals and groups embedded in data-driven algorithms lies at the core of fairness in machine learning (fair ML) research. Discriminatory practices take on various forms, affect a multitude of social groups in different contexts, and are often targeted against (intersecting) minority populations. Investigating discrimination in sociotechnical systems requires adequate and nuanced data sources as well as careful operationalizations of vulnerable groups. Data is highly influential in fair ML research. On the one hand, novel fairness methodology is typically developed and ``benchmarked'' in empirical applications, and thus the underlying data can be used to support the argument in favor of a specific technique. On the other hand, the information that is encoded and readily accessible in fairness data defines the scope of what can be tested empirically, priming fairness research to e.g. focus on those protected attributes that are most easily accessible. Practices concerning \emph{which} data is used in published research, and \emph{how} it is used, further set a standard for both practitioners and future research.

In this work, we study data practices in fairness research and identify common shortcuts that undermine its reach and reliability. Particularly, we study which protected groups are represented in datasets commonly used in fair ML and how the available data is utilized in the literature, identifying blindspots such as neglected identities and omitted subpopulations in data usage. We argue that through their wide range of applications, fairness datasets and their uses play a pivotal role in fairness research as they can be both drivers and barriers for sound methodological and empirical research.

More specifically, we study the \emph{content} of fairness datasets in interaction with their \emph{uses} in empirical research. This dual view is motivated by the concern that limitations inherent to the datasets themselves can be exacerbated by unreflective choices made in the processing and handling of these data. Both factors can jointly accumulate to the risk of neglecting ``uncommon'' protected attributes or specific subpopulations and contribute to normalize this practice, leading to a vicious cycle of canonical fairness research which focuses on a limited set of social groups and the same standard datasets \citep{fabris2022algorithmic}.

\namedpar{Related work}. Critical studies have challenged research practices in fair ML on various grounds. Concerns have been raised regarding its narrow and too granular focus, tendencies of insularity \citep{laufer2022four}, inconsistent notions of race \citep{abdu2023empirical}, and a predominance of shallow discussions of specific negative impacts that neglect structural and social factors \citep{birhane2022forgotten}. Critical data studies \citep{iliadis2016critical,boyd2012critical} view these questions from a data-centric lens. Selected challenges have been tied to the empirical foundation of fair ML research, such as its overreliance on WEIRD (Western, Educated, Industrialized, Rich, and Democratic) samples \citep{septiandri2023weird} and a large share of fairness publications drawing on the same datasets, namely Adult, COMPAS and German Credit \citep{fabris2022algorithmic}. As these data come with considerable limitations \citep{bao2022it, ding2021retiring}, there is a risk of self-perpetuating practices that steer empirical fairness research away from the social realities and diversity its data is supposed to represent.

\namedpar{Contributions}. Against this background, we focus on both the scope of fairness datasets and their uses in empirical research to understand the interaction between limitations in datasets and the choices that are made in the handling of these data. We study 280 experiments across 142 fair ML publications and identify gaps in collective data practices hindering the reach and reliability of the field. Our study makes the following contributions:

\begin{itemize}
    \item We present an inclusive list of attributes protected by anti-discrimination legislation across multiple continents and study their (under)representation in fairness datasets, as well as discrepancies between protected attribute availability and usage in fair ML research.
    \item We outline exclusionary patterns in empirical studies and demonstrate how a lack of transparency and unreflective processing choices normalize the omission of minorities and lead to ambiguous results in fairness research. 
    \item We provide actionable recommendations to remedy existing limitations and pave a path forward towards more thoughtful and nuanced data practices in fair ML.
\end{itemize}

We start by outlining our selection and annotation process of fairness datasets and publications in Section~\ref{sec:methods}. 
In Section~\ref{sec:neglected}, we contrast the availability and usage of protected attributes in fairness data with the salience of protected attributes in legislation across the globe. In Section~\ref{sec:omitted}, we demonstrate exclusionary data practices against minorities with a case study on COMPAS data. In Section~\ref{sec:opaque}, we focus on transparency and generalization, showing opaque design decisions affecting fairness evaluations with a second case study on the Bank dataset. We summarize our findings in Section~\ref{sec:discussion}, providing a list of recommendations towards better data practices in Section~\ref{sec:recs}, and concluding remarks in Section~\ref{sec:conclusion}.

\section{Methodology} \label{sec:methods}

For this work, we collected and annotated tabular dataset usage for fair classification tasks. To create this corpus, we built on top of a comprehensive survey of fairness datasets \citep{fabris2022algorithmic}, leveraging the same inclusion criteria for publications. We focus on tabular datasets and fair classification for their prominent role in the fairness literature \citep{fabris2022algorithmic,fabris2022tackling,mehrabi2022survey}. We study the use of tabular datasets ($N=36$) across 142 articles. Since many datasets appear in multiple publications and most publications use multiple datasets, the total number of dataset and publication combinations annotated was $N = 280$. 

Information regarding the usage of different datasets was collected for each combination of dataset and publication. This information includes which variant of a dataset was used, which attributes were considered protected and whether sufficient information was available to reconstruct this, as well as the target variable and features used for prediction. To collect this information, the publications, their supplementary materials, and appendices were consulted for information regarding each dataset usage. Moreover, each publication was searched for mentions of source code; if unsuccessful, we searched on the internet for code repositories mentioning the publication's title. Detailed information on the annotation process and corpus selection is available in Appendix~\ref{annotation-process}.

The collected data on dataset usage as well as the code for all analyses presented in this work are publicly available at \url{https://github.com/reliable-ai/lazy-data-practices}. Analyses were conducted and visualizations created using Python version 3.9 \citep{van1995python}, R version 4.2.2 \citep{rcoreteam2022r} and RawGraphs version 2.0 \citep{mauri2017rawgraphs}.

\section{Neglected Identities} \label{sec:neglected}

Acknowledging the diversity of vulnerability in fair ML is critical as the social impacts of prediction algorithms and the effectiveness of bias mitigation strategies can vary greatly between different protected groups. Vulnerable identities will not benefit from fairness research unless explicitly considered by it. This section studies the availability and usage of protected attributes in fair ML, which we introduce in the following subsections and summarize in Figure \ref{fig:sensitive-attributes}.

\subsection{Protected Attributes Globally}
To define protected attributes, we draw from domain-specific legislation and human rights law. We define as \emph{protected} all socially salient attributes explicitly mentioned as prohibited drivers of discrimination and inequality. For example, Article 2 of the Universal Declaration of Human Rights states ``Everyone is entitled to all the rights and freedoms set forth in this Declaration, without distinction of any kind, such as race, colour, sex, language, religion, political or other opinion, national or social origin, property, birth or other status'' \citep{un1948united}. 

On the one hand, we try to mitigate the \emph{Global North bias} in AI ethics research \citep{okolo2022making,roche2021artificial,septiandri2023weird} by covering international human rights instruments from around the globe, including the Universal Declaration of Human Rights \citep{un1948united}, the African Charter on Human and Peoples’ Rights \citep{oau1981african}, the  Arab Charter on Human Rights \citep{council2004arab}, the ASEAN Declaration of Human Rights \citep{asean2009association}, the American Declaration of the Rights and Duties of Man \citep{oas1948american}, and the Charter of Fundamental Rights of the European Union \citep{eu2000charter}.
On the other hand, we align with this bias, including a regional perspective on anti-discrimination in hiring and lending based on US and EU legislation \citep{fabris2023fairness,chen2019fairness}, covering, for example, the Fair Housing Act \citep{uscongress1968fha}, the Equal Credit Opportunity Act \citep{uscongress1974ecoa}, the Racial Equality Directive \citep{council2000council}, and the Employment Equality Directive \citep{council2000council2}. There are two mutually reinforcing reasons for this, namely the convenient availability of summary articles on the topic and the influence of these regions on anti-discrimination and fairness research.

\definecolor{lightgray}{gray}{0.95}
\begin{table}[]
\caption{\textbf{Protected attributes in global anti-discrimination law}. Protected attributes are found in international human rights instruments and domain-specific anti-discrimination law. We report a tick ($\checkmark$) when the literal phrasing (in the original law or in official clarifications) matches the row header. We report the literal phrasing otherwise.}\label{tab:sens}
\footnotesize
\rowcolors{1}{white}{lightgray}
\begin{tabular}{p{2.4cm}p{1.2cm}p{1.2cm}p{1.2cm}p{1.2cm}p{1.2cm}p{1.2cm}p{1.2cm}p{1.2cm}}
\hiderowcolors
      & \textbf{UN \mbox{Charter} \citep{un1948united}}                        & \textbf{African Charter \citep{oau1981african}} & \textbf{Arab \mbox{Charter} \citep{council2004arab}} & \textbf{ASEAN Declaration \citep{asean2009association}} & \textbf{American Declaration \citep{oas1948american}} & \textbf{EU \mbox{Charter} \citep{eu2000charter}}       & \textbf{US Fair Lending \citep{chen2019fairness}} & \textbf{EU Fair Hiring \citep{fabris2023fairness}} \\
\hline
\showrowcolors
\multicolumn{9}{l}{\textit{Gender and Sexual Identity}}                                                                                                                                                                                                                                                \\
\hspace{0.2cm}Sex                               & \tick                                                & \tick                           & \tick   &                                    & \tick                                   & \tick                               & \tick                              & \tick                             \\
\hspace{0.2cm}Sexual orientation                &                                                    &                                  &    &                                    &                                       & \tick                &                               \tick   & \tick              \\
\hspace{0.2cm}Gender                            &                                                    &                                  &    & \tick                             &                                       &                                   & Gender identity                  & Gender; gender reassignment     \\
\hline
\multicolumn{9}{l}{\textit{Racial and Ethnic Origin}}                                                                                                                                                                                                                                                  \\
\hspace{0.2cm}Race                              & \tick                                               & \tick                           &  \tick    & \tick                               & \tick                                  & \tick                              & \tick                             & Racial \mbox{origin}                   \\
\hspace{0.2cm}Color                             & \tick                                             & \tick                           &   \tick  &                                    &                                       & \tick                            & \tick                            &                                 \\
\hspace{0.2cm}Ethnic origin                         &    \mbox{Territory to} which person belongs              & Ethnic group                   &      &                                    &                                       & \tick                     &                                  & \tick                   \\
\hspace{0.2cm}National origin                       & \tick & \tick              &     \tick & \tick                    &                                       &                 Nationality                  & \tick                  &                                 \\
\hspace{0.2cm}Language                          & \tick                                           & \tick                     &  \tick  & \tick                           & \tick                              & \tick                          &                                  &                                 \\
\hspace{0.2cm}National minority &                                                    &                                  &       &                              &                                       & \tick &                                  &                                 \\
\hline
\multicolumn{9}{l}{\textit{Socioeconomic Status}}                                                                                                                                                                                                                                                     \\
\hspace{0.2cm}Social origin                     & \tick                                      & \tick               &   \tick  & \tick                      &                                       & \tick                     &                                  &                                 \\
\hspace{0.2cm}Property                          & \tick                                           & Fortune                    &  Wealth    & Economic status                    &                                       & \tick                          &                                  &                                 \\
\hspace{0.2cm}Recipient of public \mbox{\hspace{0.2cm}assistance}    &                                                    &                         &         &                                    &                                       &                                   & \tick   &                                 \\
\hline
\multicolumn{9}{l}{\textit{Religion, Belief and Opinion}}                                                                                                                                                                                                                                              \\
\hspace{0.2cm}Religion                          & \tick                                           & \tick                    &  Religious belief   & \tick                           & Creed                                 & Religion or belief                & \tick                         & Religion or belief              \\
\hspace{0.2cm}Political opinion                 & \tick                                  & \tick                &   & \tick         &                                       & \tick  &                                  &                                 \\
\hspace{0.2cm}Other opinion                     & \tick                                      & \tick                &         Opinion; thought        &  \tick                                  &                                       & \tick  &                                  &                                 \\
\hline
\multicolumn{9}{l}{\textit{Family}}                                                                                                                                                                                                                                                                    \\
\hspace{0.2cm}Birth                      & Birth status                                       & Birth status                 &   \tick  & \tick                              &                                       & \tick                             &                                  &                                 \\
\hspace{0.2cm}Familial status                   &                                                    &                          &        &                                    &                                       &                                   & \tick                  &                                 \\
\hspace{0.2cm}Marital status                    &                                                    &                          &        &                                    &                                       &                                   & \tick                   &                                 \\
\hline
\multicolumn{9}{l}{\textit{Disability and Health Conditions}}                                                                                                                                                                                                                                                   \\
\hspace{0.2cm}Disability                        &                                                    &                          &    \tick    & \tick                         &                                       & \tick                        & \tick                       & \tick                      \\
\hspace{0.2cm}Genetic features                  &                                                    &                          &        &                                    &                                       & \tick                  &                                  &                                 \\
\hline
\multicolumn{9}{l}{\textit{Age}}                                                                                                                                                                                                                                                                       \\
\hspace{0.2cm}Age                               &                                                    &                           &       & \tick                                &                                       & \tick                               & \tick                              & \tick                             \\
\end{tabular}
\end{table}

Drawing from this literature, we provide a shallow categorization of protected attributes, reported in Table \ref{tab:sens}. We identify seven main categories for protected attributes: (1) gender and sexual identity, (2) racial and ethnic origin, (3) socioeconomic status, (4) religion, belief and opinion, (5) family, (6) disability and health conditions, and (7) age. Most protected attributes fall into at least one of these categories. We categorize attributes potentially relevant to more than one category, such as ``genetic features'', based on specialized literature \citep{de2015regulating}. It is worth noting \textbf{this is not a complete categorization} of all protected attributes around the globe and across sectors.\footnote{For example, veteran status does not appear in Table \ref{tab:sens}, despite being protected in certain countries and industry sectors. Moreover, we neglected the right to non-discrimination for exercising CCPA rights under the California Consumer Privacy Act \citep{chen2019fairness} since it applies in a single country.} This categorization aims to guide an inclusive discussion of algorithmic fairness research through the lens of protected attributes. 
 
\subsection{Who is Missing}

\begin{figure}%
    \includegraphics[width=0.95\textwidth]{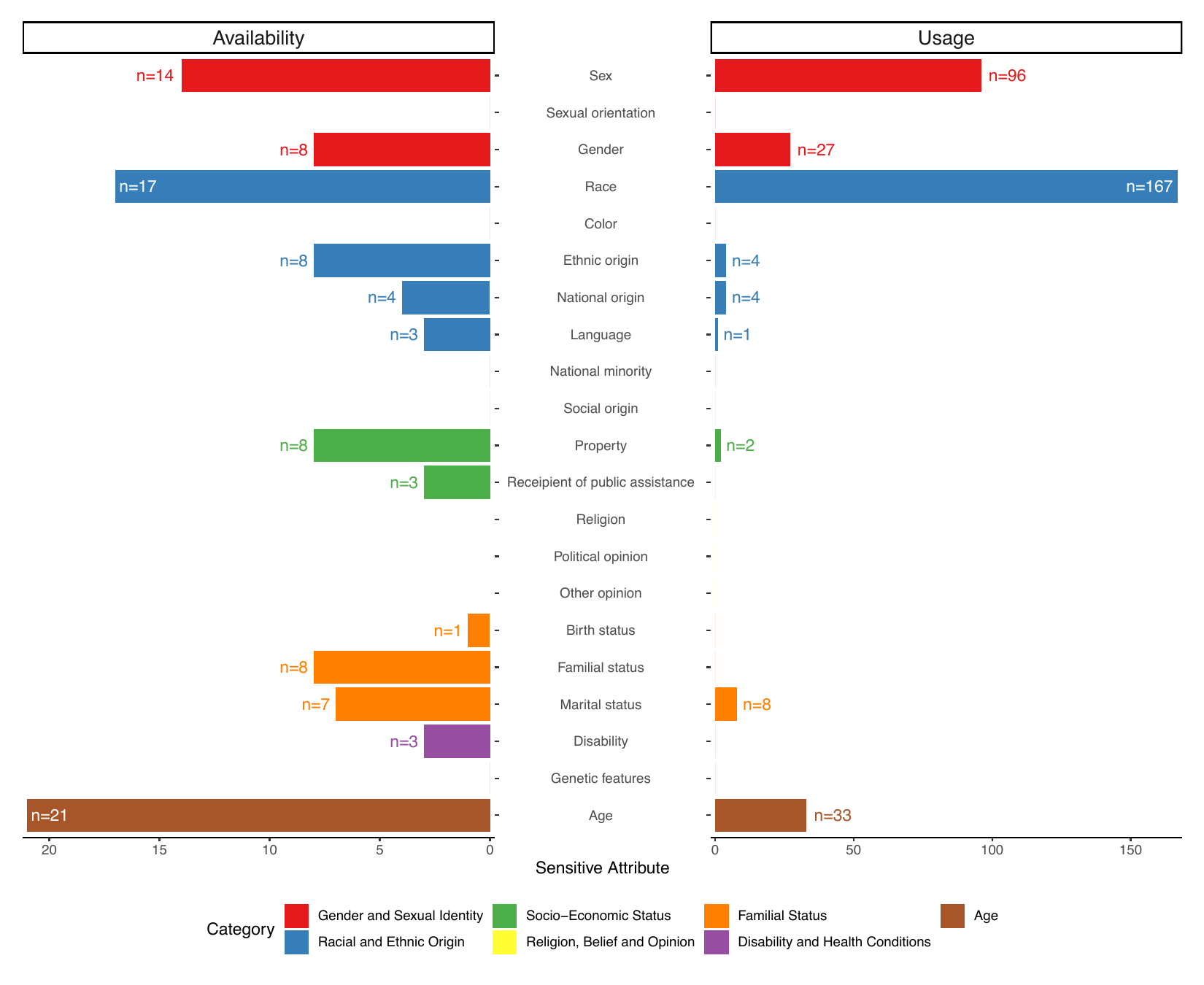} 
    \caption{\textbf{There is a large discrepancy between the list of attributes considered protected under international legislation and their availability or usage in datasets}. Bar chart displaying the availability (left) and usage (right) of protected attributes in the literature for all categories of protected attributes in Table~\ref{tab:sens}. Availability based on a total of $N = 36$ datasets; usage based on a total of $N = 233$ experiments with enough information available to reconstruct (or at least make an educated guess about) protected attribute usage (see Section~\ref{sec:opaque} regarding a lack of available information).}%
    \label{fig:sensitive-attributes}%
\end{figure}

Incentives against the collection and use of protected data are well documented in the literature \citep{andrus2021what}, motivating the line of work on fairness under unawareness \citep{chen2019fairness,fabris2023measuring}, which aims to measure and improve fairness with no access to protected attributes. In this section, we demonstrate that this effect is not uniform across all protected attributes. The left bar chart in Figure~\ref{fig:sensitive-attributes} depicts protected attributes available in popular fairness datasets. Attributes about \emph{religion, belief and opinion} are entirely missing. Variables describing \emph{disability and health conditions} are very infrequent ($n=3$) and never used in the surveyed literature (right bar chart in Figure~\ref{fig:sensitive-attributes}). \emph{Socioeconomic status} descriptors are more commonly available yet frequently neglected.\footnote{For completeness, we also encountered a small number of protected attributes used in the literature but not referenced in legislation, including employment status, alcohol consumption, neighborhood, body-mass index, and profession.}

Some protected attributes are particularly sensitive and safeguarded by data protection law. The GDPR (General Data Protection Regulation \citep{european2016gdpr}) bans the use of special categories of personal data, including religion and health data, making it more difficult to collect and use these data to audit or train algorithmic systems \citep{van2023using}. The Americans with Disabilities Act \citep{unitedstates1990americans} imposes strict regulations to disability-related questions that employers can ask \citep{uscongress1990ada}. Data protection, however, does not fully explain the availability and usage of protected attributes in fairness research. In the following, we detail the causes and effects of neglecting protected identities.

\namedpar{Disability} is a highly diverse, nuanced, and dynamic construct \citep{trewin2018ai}. Technological ableism is pervasive \citep{shew2020ableism}; algorithmic fairness is insufficient to counter it as it tends to oversimplify and flatten disability. Indeed, there have been multiple calls to move beyond simplistic notions of fairness and towards disability justice \citep{bennett2020what,tilmes2022disability}. However, this fundamental recognition of nuance may act as a double-edged sword. Even in specific contexts where disability can be treated more narrowly, such as speech recognition for people with speech disorders, data is sparsely available \citep{papakyriakopoulos2023augmented}. Research highlighting biases across speech impairments \citep{green2021automatic,hidalgo2023quantifying} has not gained traction in algorithmic fairness venues \citep{trewin2019considerations,buyl2022tackling}. Overall, it seems plausible that other protected attributes have been prioritized, to the detriment of disabled identities, due to difficulties in handling a diverse spectrum of conditions, complex data ethics, and concerns of oversimplification. Acknowledging its limitations, we believe that fair ML research can benefit people with disabilities, especially for bias detection and analyses of its root causes.

\namedpar{Religion} and creed are protected by all surveyed legislations. They are a strong driver of identity, bias, and prejudice; in the extreme, they can lead to violence \citep{chuah2016religion,amarasingam2022fight}. Religion is highly salient in specific contexts, for example materializing as anti-Muslim discrimination in Western societies \citep{ahmed2010muslim,fernandez2023discrimination,abid2021persistent}. Data collection, however, remains contingent on political will \citep{sambasivan2021reimagining,ghumman2013religious}. It is often unavailable in census data \citep{gutierrez2017census,uscensus2022does} and laws mandating data collection for anti-discrimination, such as the HMDA (Home Mortgage Disclosure Act \citep{uscongress1975hmda}), do not include religion \citep{andrus2022demographic}. Indeed the effectiveness of Western anti-discrimination law in protecting religious minorities such as Muslim identities has been called into question \citep{bloul2008anti}. Negative stereotypes of Muslims have been documented in different regions of the world \citep{sides2013stereotypes,brown2017us,van2019muslim}. While fairness research has been able to study Muslim bias in language models \citep{abid2021persistent,muralidhar2021examining,dhamala2021bold}, so far it has neglected allocative harms against Muslim people. It could be argued that a lack of focus on religion is compensated by research on racial and ethnic discrimination, since religions have strong ethnic foundations, and congregations tend to be racially homogenous \citep{kim2011religion,chaves1998national}. However, religious and ethnic discrimination can compound rather than simply overlap \citep{distasio2021muslim}. Moreover, racial classifications are insufficient for Middle Eastern and North African people, who are classified as white by the US government \citep{maghbouleh2022middle}. Overall, fairness research has neglected this important axis of discrimination and its intersections with other vulnerable identities \citep{nadal2015qualitative,fernandez2023discrimination,sambasivan2021reimagining}.

\namedpar{Property}. 
High-tech tools can disempower poor people \citep{eubanks2018automating,kirkpatrick2021algorithmic}. Stakeholders of child protection systems are concerned about models automating biases against the poor \citep{stapleton2022imagining}. Overall, poverty shows mutually reinforcing negative effects on health, education, and justice \citep{fuller2012poverty,parolin2022role,ladd2012education,rabuy2016detaining}. Despite this fact, property and other socioeconomic variables are seldom used as protected attributes in algorithmic fairness research. This is partly due to data availability: poverty data from household surveys is coarse and sometimes unavailable, especially in the developing world \citep{noriega2020algorithmic}. In addition, and perhaps to a greater extent, it is due to data usage. Wealth is often the target variable of models, such as algorithmic social policies \citep{noriega2020algorithmic,hanna2018universal}, or one of their (unprotected) input features, as in creditworthiness estimators \citep{das2023algorithmic}.  This seems especially true in fairness research, where the most popular task is income prediction with the Adult dataset \citep{fabris2022algorithmic}. Among formally protected attributes, property is uniquely associated with a perception of mutability and merit: people tend to associate wealth and poverty with individual merit rather than structural constraints \citep{heiserman2017higher,bucca2016merit}. This perception fuels the discourse on deservingness, seeking to distinguish between deserving and undeserving poor people, which determines the boundaries of admissible redistribution policies \citep{applebaum2001influence,watkins2016discourse}. In turn, this impacts algorithmic fairness research, not only discouraging bias mitigation based on wealth, but also constraining measurement along this protected axis.

This section highlights blindspots in fairness research, neglecting vulnerable and globally salient identities. It is worth noting that this trend extends to fairness research more broadly, including qualitative studies, and to more protected attributes, including sexual orientation. As a prevalent practice in the field, it has a tendency to self-reinforcement, further incentivizing future research to conform. Indeed recent articles published at fairness conferences, such as \textit{FAccT} (the ACM Conference on Fairness, Accountability, and Transparency) and \textit{AIES} (the AAAI/ACM Conference on Artificial Intelligence, Ethics and Society), mention race and gender more frequently (by one order of magnitude) than religion, disability, socioeconomic status, and sexual orientation \citep{birhane2022forgotten}. Taking stock of a complex social, legal, and technical landscape, we argue for a move towards an ambitious research roadmap to tackle this complexity (as advocated, for example, in \citet{guo2020toward}); avoiding it will only prevent us from noticing and remedying existing harms. 

\section{Omitted Populations}\label{sec:omitted}

A lack of accurate and proper representation is at the heart of many issues the fairness community tries to address. Oftentimes minority groups are neglected in data, leading to discriminatory behavior of systems leveraging this data \citep{mehrabi2022survey}. Neglect is nuanced and takes many forms. It can materialize as a lack of consideration for specific protected attributes, as discussed in the previous section. It can also derive from the underrepresentation of certain groups in the population during data collection, who are not easy to reach. As we will demonstrate in this section, the issue of underrepresentation gets exacerbated due to the common practice of excluding information about smaller groups during data processing. This is often done out of convenience, to turn a multi-group problem into a binary one, or in some cases, for privacy reasons. In tabular data, this exclusion can either take the form of outright removal of minority groups from the data or aggregation of multiple minority groups into one big ``other'' group.

These exclusionary data practices are surprisingly common in the examined literature and even more concerning is that they often apply to protected attributes. As protected attributes are, by definition, linked to vulnerability, this amounts to discarding data for disadvantaged minorities. Normalizing these practices sets a dangerous example and incentive for the adoption of such practices also outside of research within real-world systems, with great potential for harm, especially to the most vulnerable populations.

\subsection*{Case Study: Omitted Identities in COMPAS}

To demonstrate this practice, we study the different processing strategies in publications using the COMPAS dataset \citep{angwin2016machine}, one of the most popular datasets in the fairness literature \citep{fabris2022algorithmic}. The Correctional Offender Management Profiling for Alternative Sanctions (COMPAS) system is a risk assessment tool used in the US judicial system. The dataset, distributed under the same acronym, was constructed by ProPublica as part of a publication describing racial biases in the profiling system. It contains risk scores from the system for individuals in Broward County, Florida, US, generated during 2013--14. A datasheet \citep{gebru2021datasheets} for the COMPAS dataset is available in the Appendix of \citet{fabris2022algorithmic}. The attribute typically considered protected is \textit{race} with a total of 6 categories: ``African-American'', ``Asian``, ``Caucasian'', ``Hispanic'', ``Native American'' and ``Other''.

\begin{figure}%
    \centering
    \includegraphics[width=1.0\textwidth]{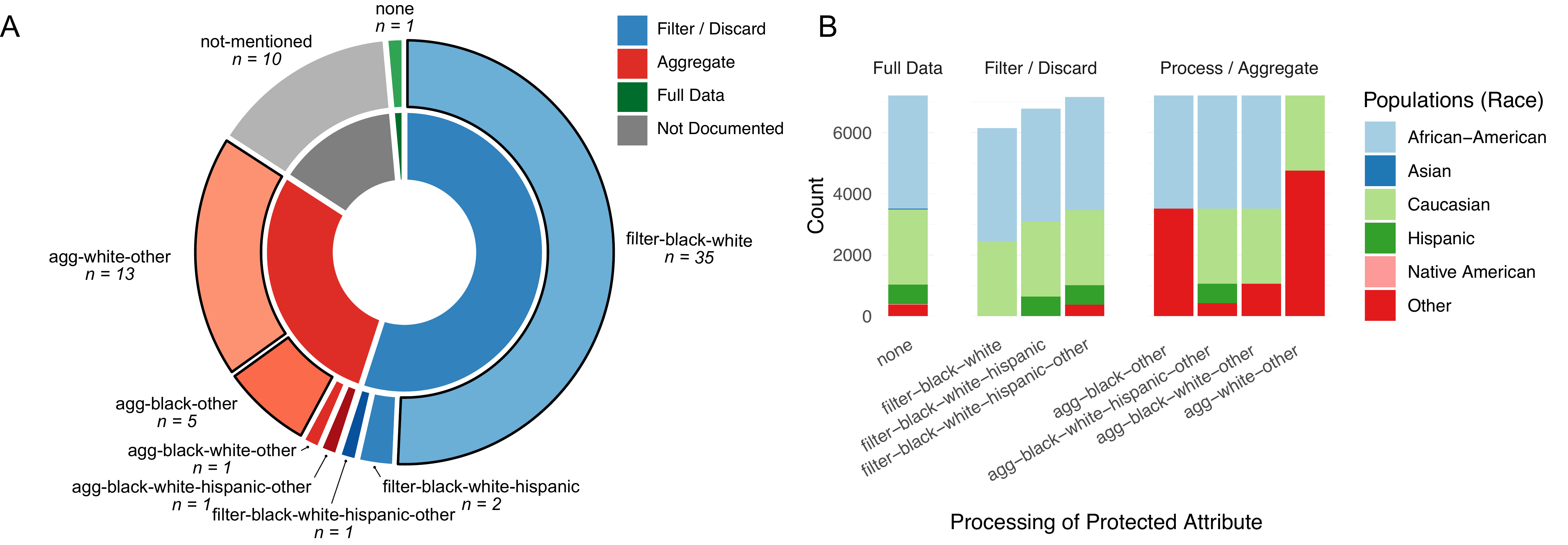} 
    \caption{\textbf{Data from smaller populations is almost \textit{always} either discarded or aggregated within the annotated literature.} (A) Prevalence of processing strategies for the COMPAS dataset within the annotated literature and (B) resulting base rates of the protected attribute from these different processing strategies. Due to the small sample sizes, the populations of Asians and Native Americans are difficult / impossible to see in the figure. Neither group is included as a category in any of the processing strategies except when using the \textit{Full Data} ($n=1$). Processing strategies binarising protected attributes (i.e. leaving a binary variable with only two groups) are highlighted with a black outline in A. The inner circle corresponds to the combined prevalence of processing strategies using a specific approach (e.g. filtering or aggregation).}%
    \label{fig:processing}%
\end{figure} 

Overall, we annotate $N = 69$ publications using the COMPAS dataset, with $85.5\%$ (59)  providing enough information to reconstruct whether and how the \textit{race} attribute was processed. Although some publications considered additional attributes to be protected, we did not systematically annotate processing of other protected attributes.
We identify a total of 8 different processing strategies with the frequency of their occurrence shown in Figure \ref{fig:processing}A. We sort processing strategies into three categories: (1) \textit{none} if all data was retained as-is, (2) \textit{aggregating} if all observations were retained, but subgroups were recoded and aggregated e.g. collapsing data into ``African-American'' and ``Other'', and (3) \textit{filtering} if observations were discarded rather than recoded or aggregated, e.g. keeping only the groups ``African-American'' and ``Caucasian'' (the most common form of processing). We do not observe a combination of aggregating and filtering, although such a strategy could easily be conceived. Examining Figure \ref{fig:processing}A, we see that only a single publication examined the full data as-is. The overwhelming majority of publications either filter/discard (38) or aggregate (20) populations. The most extreme processing strategies, leaving only two groups, are the most common (53).

To highlight how processing strategies affect data, we apply each processing strategy on the COMPAS dataset and show the distribution of the resulting \textit{race} attribute in Figure \ref{fig:processing}B. While we compare all processing strategies on the same version of the COMPAS dataset (compas-scores-two-years.csv), we observe different publications using different versions of the dataset. Figure \ref{fig:processing} demonstrates how different strategies for data processing alter the composition and distribution of protected attributes. Many of the strategies leave only two groups, either discarding or aggregating minority groups; none of the actual processing strategies retain Asian or Native American populations as distinct groups. In general, few papers describe, and even fewer justify their choices when handling protected attributes \citep{abdu2023empirical}. 

This fact shows a tendency to simplification and binarization in fair ML empirical research, which seems at odds with the importance of diversity and socio-technical context broadly acknowledged in this field.  We speculate that this is partly driven by methodological advances which are more practical under binary protected attributes, and partly by a tendency to algorithmic benchmarking, which is more straightforward in the binary setting. Binarization as an implicit norm in the literature sets a dangerous precedent for research and practice in the field. As a consequence, we see a risk of omission disproportionately affecting vulnerable minorities. Besides the dangerous precedent of normalizing the exclusion of vulnerable subgroups from the data, this also threatens the transparency and reproducibility of fairness research; Figure \ref{fig:processing}A demonstrates a large share of publications without enough information to reconstruct processing decisions. It is worth noting that, while different publications use different versions of the dataset, this section focuses on a single dataset for comparability and simplicity. Our results, therefore, give a lower bound on data processing variation. As the next section shows, these opaque and diverse choices can lead to very different outcomes during model evaluation and comparison. 

\section{Opaque Preprocessing}\label{sec:opaque}

The previous section describes disparate practices for protected attribute processing that are often overlooked. This section discusses a broader lack of documentation on dataset usage and its consequences. This is a significant risk to the reproducibility and generalization of fairness research for a combination of two reasons: (1) many publications do not document their usage of a dataset sufficiently, assuming that merely the name of a dataset clearly identifies its usage and (2) publications that do document data usage or offer reproducible code vary greatly in their usage, disproving the idea that merely identifying a dataset by its name is sufficient information. These variations in usage, or preprocessing, are likely to affect fairness \citep{mougan2023fairness,simson2023everything}.
Beyond the variation in the mere \textit{usage} and processing of a dataset, we also observe many publications using different \textit{variants} or versions of datasets, sometimes from the same official source and sometimes from undocumented sources. These variants often lack information regarding the processing that happened to create them.

For each dataset-publication combination experimenting with a prediction task ($N=262$),\footnote{18 publications do not fit the typical paradigm of using features to predict a target variable and are therefore omitted. Experiments on synthetically generated datasets are coded as \emph{Not Applicable}.} we annotated the level of documentation, including whether a publication included enough information to reconstruct dataset usage. In particular, we annotated the level of information regarding (1) the target variable that was being predicted $y$, (2) the features used for classification $X$, and (3) the protected attributes $S$. We graded each publication for each aspect into one of three levels: \textit{Yes}, if there was sufficient information, \textit{Guessable} if someone familiar with the dataset could reasonably make an educated guess, and \textit{No} if there was insufficient information or none at all provided. For each publication, we looked for information in the main publication, the supplementary materials, and the source code. We annotated the availability of source code for every dataset-publication pair ($N = 280$). As source code was often not directly referenced in publications, we also searched for it explicitly for every annotated experiment. If source code was available with a certain publication but did not match the publication's analyses, we discarded it as \emph{Not Available}. An example of this are articles presenting new methodologies and experiments, which provide an implementation of the new method but no code reproducing their experiments.

\begin{figure}%
    \centering
    \includegraphics[width=0.6\textwidth]{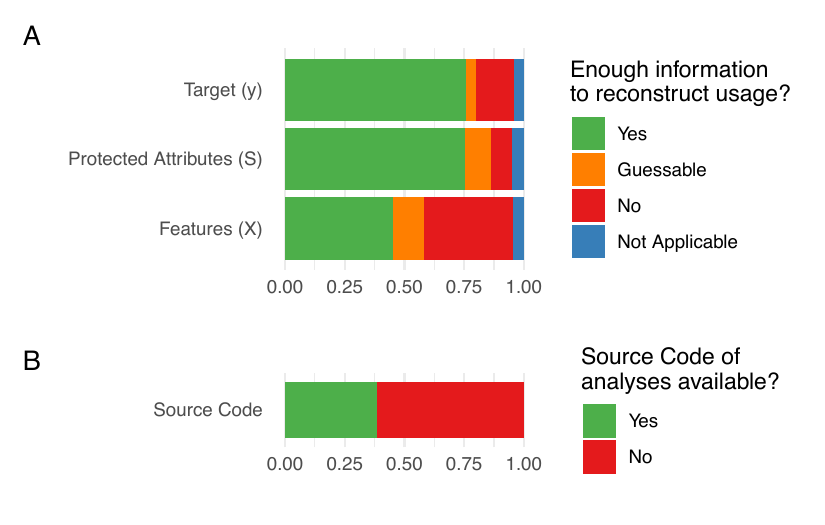} 
    \caption{\textbf{A large section of the annotated literature lacks sufficient information to reproduce analyses.} Bar diagrams showing whether publications in the annotated literature contain (A) sufficient information to reconstruct usage of the predicted target variables $y$, the protected features $S$ and the features used for prediction $X$ and (B) source code to reproduce analyses. Only publications containing a prediction task are included in the figure.}%
    \label{fig:info-quality}%
\end{figure} 

The resulting annotations are summarized in Figure \ref{fig:info-quality}, showing that the provided information was insufficient to reconstruct the target variable for 16\% (41 out of 262) of annotated experiments and 9\% (23) of experiments were lacking information regarding protected attributes. Regarding features, the situation is even worse, with \textit{half} of the annotated experiments (132) containing either not enough information (98) or forcing one to guess (34) to reconstruct feature usage. As publications themselves seldom provide sufficient information to reconstruct dataset usage, this issue is also largely due to a lack of available source code, with just 39\% (108 out of 280) of publications providing source code for their analyses. This lack of documentation is problematic for both the reproducibility of research and the generalization of findings in the field, as we will demonstrate in the following.

It is worth noting that proper documentation of preprocessing choices is not sufficient on its own. For example, 10 out of 22 publications using the ``German Credit'' dataset report extracting $gender$ or $sex$ information from the data. This is based on the widespread misbelief that this information can be extracted from a column in the dataset, when in fact the necessary information is not available \citep{groemping2019south}. Nonetheless, having this information explicitly available in the respective publications allows readers to evaluate essential aspects of their correctness and quality. 

\subsection*{Case Study: Opaque Preprocessing of Bank}
We demonstrate the extent and impact of the variation in dataset usage using the ``Bank Marketing'' dataset \citep{moro2014data} (from here onwards: Bank). This dataset is quite relevant in fairness research (fifth most popular \citep{fabris2022algorithmic}) yet understudied in the literature. Bank describes telemarketing of long-term deposits at a Portuguese bank in the late 2010s. Instances represent telemarketing phone calls and include client-specific features (e.g. job and age), call-specific features (e.g. duration), and environmental features (e.g. euribor). The associated task is to predict whether clients subscribed to a term deposit after the call.

\subsubsection*{Disparate Preprocessing Choices}

We compiled a short list of structured preprocessing choices for Bank across 9 scholarly articles in our corpus focusing on dataset version and protected attributes. First, we note which version of the dataset was used, as there are a total of four different versions available in the original source, two of which have been used in our corpus: \textit{bank-full} and \textit{bank-additional-full}, with the version marked as \textit{additional} containing additional variables, but having slightly fewer observations than the other version. Second, we examine which attributes were considered protected, and third, how they were processed. 

\begin{figure}%
    \includegraphics[width=0.8\textwidth]{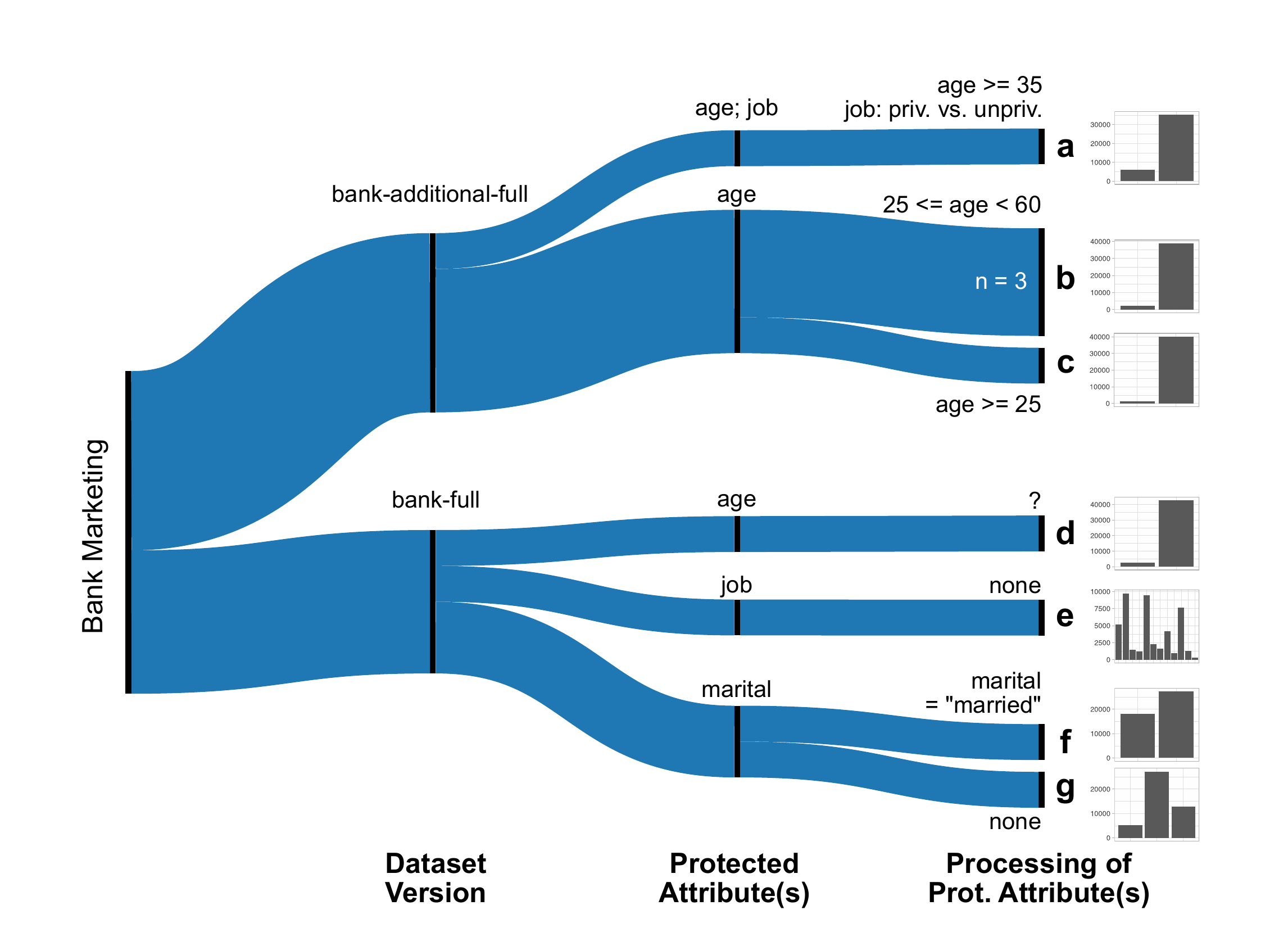} 
    \caption{\textbf{The ``same'' dataset is used in many different ways within the literature.} Sankey diagram illustrating the usage of the Bank dataset within the annotated literature. Each split corresponds to a choice where differences were observed in the literature. Each unique combination of choices or scenario is identified by a unique letter, with the base rates of the protected attribute(s) displayed on the right. We constructed this figure to provide a conservative, lower-bound estimate regarding the variation in dataset usage.}%
    \label{fig:sankey}%
\end{figure}

We find \textit{age}, \textit{job}, and \textit{marital} to be considered protected, with one publication considering both \textit{age} and \textit{job} protected. While most examined publications consider \textit{age} protected, they show variability in its preprocessing. We identify 3 different strategies to turn age into a binary column.\footnote{Strictly speaking there are 4 different strategies, as we observe a single publication processing age as ``age < 25 or age > 60'' as opposed to ``age >= 25 and age < 60'' which was observed in two other publications. As these two strategies are equivalent to each other except in how they encode individuals who are exactly of age 60, we combine them under the more popular choice. Moreover, one publication does not mention processing the protected attribute, in which case we also use the most popular processing strategy, as keeping age unprocessed would have been unrealistic.} Overall, the 9 publications produce 7 distinct combinations of these three choices. An overview of these scenarios, alongside a visualization regarding the prevalence of each choice, is presented in Figure~\ref{fig:sankey}. Notice we are not considering additional choices in dataset processing, such as selection of non-protected features ($X$), thereby providing a lower bound on the variation in the usage of Bank.

\subsubsection*{Impact of Disparate Preprocessing}

As shown in Figure \ref{fig:sankey}, disparate data processing choices translate into variations in the base rates of the protected attributes, shown beside the identifying letter of each scenario. To quantify the impact of this variation on algorithmic fairness, we consider a fair classification task with the different scenarios in Figure \ref{fig:sankey}. For each scenario, we fit and examine multiple models using the state-of-the-art automated machine learning library \emph{autogluon} version 1.0 \citep{agtabular,gijsbers2023amlb,autogluonbenchmark}. A total of $N = 13$ models are considered; 12 correspond to the default model/hyperparameter configurations in \emph{autogluon} plus a logistic regression model, included for its popularity in the literature and its common use in practice. We use the variable \textit{y} as a target, consider all non-protected columns as features, and evaluate fairness using the protected attributes as processed under each scenario. We evaluate the performance (F1 score) and fairness (equalized odds difference \citep{hardt2016equality}) of each model, averaging across 10 train-test splits. The fairness and performance measures used in this work are defined in Appendix \ref{app:opaque}. The within-scenario variations of both measures are sizeable with an average spread ($\bar{\delta} = mean(max(x) - min(x))$) of $\bar{\delta}_{EOD} = 0.10$ for equalized odds difference and $\bar{\delta}_{F1} = 0.20$ for F1 score across all scenarios and splits. 

\begin{figure}%
    \includegraphics[width=0.9\textwidth]{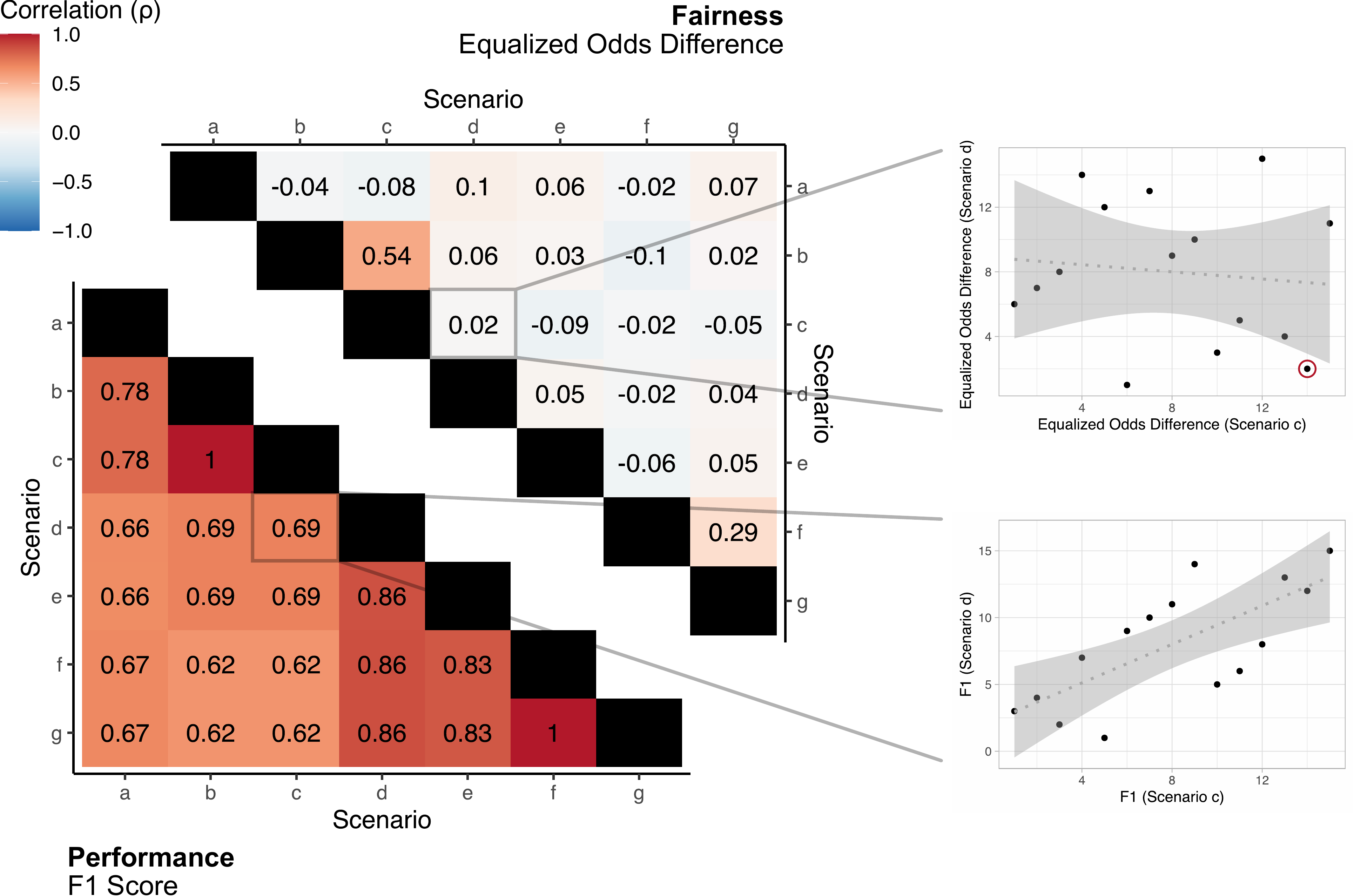} 
    \caption{\textbf{While a practitioner would choose roughly similar models based on performance across the different scenarios, they would choose very different ones based on fairness.} Spearman's $\rho$ correlations of model ranks on a measure of fairness (Equalized Odds Difference) and performance (F1 score) between different scenarios. Letters correspond to scenarios described in Figure~\ref{fig:sankey}.}%
    \label{fig:correlations}%
\end{figure} 

Within each scenario, we rank models based on their performance as well as their fairness scores, mimicking a model comparison and selection process based on accuracy and fairness evaluation. We compare model rankings across scenarios to estimate the impact of data processing choices. We compute Spearman rank correlations ($\rho$) on these rankings, reporting the full correlation matrices in Figure \ref{fig:correlations}. 

Correlations are high for performance measures (F1 score), with a mean of $\bar{\rho}_{_{F1}} = 0.747$. This means that model comparison and selection based on performance is stable and generalizes across different scenarios.  When examining correlations based on fairness, we observe significantly lower and much more variable (sometimes even negative) correlations, with a mean $\bar{\rho}_{_{EOD}} = 0.04$. This finding suggests that model comparisons based on equalized odds are strongly dependent on different data processing scenarios. The plots on the right in Figure \ref{fig:correlations} exemplify this fact, depicting model comparisons for a single run of the analysis under scenarios \texttt{c} and \texttt{d} based on F1 score (bottom) and equalized odds difference (top). A rank correlation close to zero for fairness-based rankings entails that the fairest model in scenario \texttt{c} may be among the least fair in scenario \texttt{d}. For example, the second-best model for equalized odds under scenario \texttt{c} (highlighted in red) is the second-worst performer under scenario \texttt{d}. Comparing model fairness under different data processing scenarios yields completely different results. Additional results of this analysis can be found in Appendix~\ref{app:opaque}, including correlation matrices using balanced accuracy and demographic parity \citep{calders2009building} (Figure~\ref{fig:corplot-balacc-demparity}).

Additionally, we extend our analysis to algorithms designed specifically for fair ML by training and evaluating the methods in \citet{friedler2019comparative} on the Bank data from each scenario. We used the exact same list of algorithms as the original work \citep{friedler2019comparative,calders2010three,kamishima2012fairness,feldman2015certifying,zafar2017fairness}. This experiment, reported in Appendix~\ref{app:opaque} (Figure~\ref{fig:corplot-friedler}), confirms the instability of fairness-based model comparisons under these preprocessing choices. Overall, the results demonstrate how variability in dataset usage translates into variability of fairness scores; fairness-aware experiments would choose very different models based on the different scenarios, despite working with the ``same'' Bank dataset.

\section{Discussion} \label{sec:discussion}

In the present article, we demonstrate how common choices in algorithmic fairness datasets harm the quality and curb the impact of fair ML research. We identify multiple worrying aspects regarding prevalent data practices in the literature. First, we notice that \textbf{several protected attributes are neglected} (Section \ref{sec:neglected}). This problem is partly due to privacy concerns and is exacerbated by how datasets are used in practice, with many publications focusing on a small fraction of protected attributes while relying on an even smaller number of datasets. 

Moreover, we find that \textbf{smaller subpopulations are often excluded from analyses} (Section \ref{sec:omitted}), either by aggregating all subpopulations into a single ``Other'' group or by just outright dropping their data. Therefore, rare identities, such as religious minorities or people with uncommon disabilities, have a double risk of being neglected: important protected attributes are often unavailable, and when they are, small minorities can be filtered out or aggregated for convenience. This is an exclusionary practice that fair ML work should not normalize, but rather counter. Ultimately, misrepresentation of minorities and careless processing choices have been identified as sources of biases in the first place \citep{rodolfa2020bias}, and thus represent practices that should not be reproduced by fairness research itself. We further note that neglecting minorities limits research on intersectionality as the identification of intersectional subgroups depends on the presence of (all) interacting attributes and their sufficient representation in data. 

Last, we observe a large amount of variation in the practical usage of datasets which leads to very different model comparisons based on fairness properties. Paired with the lack of proper documentation, this poses a \textbf{threat to the reproducibility and generalization of experimental results}  (Section \ref{sec:opaque}), potentially misleading practitioners during model evaluation and selection.

\namedpar{Limitations}. There are certain limitations to our results. First, work reflecting on the practices of the algorithmic fairness community should also study the industry perspective. This article focuses on fairness research since we were unable to conduct practitioner interviews or otherwise evaluate common practices in the industry. Although research differs significantly from industrial contexts, it certainly influences the prevalent methodologies and best practices in the field. Second, this work studies tabular datasets used for fair classification. We expect minor differences in the usage and availability of protected attributes in other data modalities and tasks, including e.g. the availability of skin type annotations in vision datasets \citep{buolamwini2018gender}.  Moreover, this work focuses on the corpus of publications studied in \citet{fabris2022algorithmic}, containing articles published up to and including 2021. While rather unlikely, data practices in the field may have significantly changed. We examine the robustness of our findings in Appendix \ref{app:robustness} by considering manuscripts covering different fair ML tasks and data modalities published in 2023. Our results indicate that the analyzed data practices largely remain the same, with the exception of the recently introduced and rapidly adopted Folktables datasets \citep{ding2021retiring}.

\section{Recommendations} \label{sec:recs}

The present results remain relevant and warrant addressing; we propose the following recommendations.

\namedpar{Careful inclusion of missing protected attributes in the data}. Attributes such as religion and disability are uncommon in fairness research and, more broadly, in machine learning datasets. Strong incentives against their collection include concerns about privacy and consent. We call for dedicated initiatives, including data donation campaigns and citizen science initiatives, capable of filling this gap and responsibly handling the collected data \citep{bietz2019data}. Targeted data collection initiatives are certainly difficult to undertake, as they require ethical reviews, advertisement through trusted parties, meaningful consent elicitation, and proper data infrastructures with permission systems. By making this gap more visible, we hope to incentivize new work in this direction, including methods to build semi-synthetic datasets that can be used for fairness research without compromising sensitive information of data subjects \citep{stadler2022synthetic,bhanot2021problem}.

\namedpar{Handling multiple small subgroups}. Discarding or aggregating data from protected subpopulations is a practice with a high potential for harm that should be countered, rather than normalized, especially by the fair ML community. If real-world data is complex, featuring multiple protected groups with skewed distributions, such complexity should be acknowledged and addressed directly. Pretending that these challenges do not exist by artificially making problems binary, harms the omitted populations immediately, as they are neglected in the present analysis, and in the long term by legitimizing exclusionary practices. First, we call for more explicit discussion about the practicality of proposed approaches beyond binary settings, as with works on intersectionality and rich subgroup fairness \citep{kearns2019empirical,wang2022towards}. Authors should be explicit (and reviewers demanding) about the applicability of techniques allegedly presented under a binary framing for ``notational convenience''. Second, the fact that omitted groups are always smaller points to an (often implicit) concern about the significance and stability of groupwise differences. Disaggregated analyses can be unstable for small groups; there is no easy way around this. We advocate the development of nuanced fairness evaluations for disaggregated analyses over small groups; such measures should convey information on uncertainty akin to confidence intervals and describe the statistical significance of differences.

\namedpar{Transparent data usage}. Silent subgroup omission is an example of a broader issue of opaque data processing. We call for reflection and transparency in the usage of datasets. Researchers should clearly document how and why specific datasets are chosen and, even more importantly, how they are used. Publications should document which version of a dataset is used (if there are several) and how exactly the data was processed. If the setting is a prediction task, they should mention which variables were predicted, which features were used for prediction, and which attributes were considered protected. Authors can use appendices and supplementary materials when brevity is important. Ideally, they should also provide the source code of analyses, following best practices regarding reproducibility and open research \citep{munafo2017manifesto,nosek2015promoting}. In this regard, we recommend including all the code used to preprocess data, even when preprocessed data is cached and made available, as it can be hard to reconstruct the origin of the data.

\section{Conclusion} \label{sec:conclusion}

In this work, we demonstrated common data practices in algorithmic fairness research, including the unavailability of certain protected attributes, the frequent omission of minority groups, and the lack of documentation about preprocessing choices that influence fairness evaluations despite being overlooked. These practices harm fairness research by neglecting vulnerable identities, leading to undetected harms, and by threatening the reproducibility and generalization of findings. They are currently normalized in the literature, where they set a dangerous precedent unless countered with thoughtful data choices. Data is at the core of this field; we hope the issues raised here will lead to better usage of existing datasets and inspire the careful curation of new resources.

\section*{Research Ethics and Social Impact}

\subsection*{Ethics Statement}
Our analyses hinge on a specific type of social data summarizing scholarly publications. In this context, authors of articles are data subjects whose interests should be considered and balanced against the need to keep community data practices in check. We believe that scientific critique of publicly available works is legitimate and that \emph{negative citations} are unlikely to have a sizable effect on the popularity of an article \citep{catalini2015incidence} and the livelihood of its authors. Despite these facts, we decided that criticism of individual manuscripts would not add much utility to our work, while potentially leading to (limited) negative consequences for their authors. Therefore, we focused on aggregate analyses of data practices without singling out individual manuscripts.

\subsection*{Positionality Statement}
All authors are affiliated with European organizations from Western, Educated, Industrialized, Rich, and Democratic (WEIRD) countries, in line with a documented pattern in this research community \citep{septiandri2023weird}. We found this bias especially relevant when sourcing definitions of protected attributes, as we were initially more inclined to consult resources representing European and North American points of view. We tried to mitigate this bias by consulting international human rights declarations and conventions from around the globe, but our background and the prevalent points of view in the research community inevitably influenced this work. 

\subsection*{Adverse Impact Statement}
Our adverse impact concerns are threefold. First, we would like to reiterate that our categorization of protected attributes in Section \ref{sec:neglected} is incomplete and partial. We are unaware of other manuscripts providing a list of globally protected attributes and therefore caution readers against considering our work a comprehensive resource on the topic. Second, our call for transparent data usage, in Section \ref{sec:recs}, implies an additional documentation effort by researchers; we believe this individual effort will benefit the research community, leading to more careful and reflective data practices as well as more reliable findings. Third, we highlight the word \textbf{careful} in our recommendation to include missing protected attributes: the tension between fairness research and data protection is especially relevant for this problem and requires careful consideration; the former should not carelessly trump the latter.

\begin{acks}

We would like to thank F. Weber and A. Kreider for their help in the annotation process.

\subsection*{Funding}

This work is supported by the DAAD programme Konrad Zuse Schools of Excellence in Artificial Intelligence, sponsored by the Federal Ministry of Education and Research, the Munich Center for Machine Learning and the Federal Statistical Office of Germany. The work by A.F. is supported by the FINDHR project, Horizon Europe grant agreement ID: 101070212 and by the Alexander von Humboldt Foundation.

\end{acks}

\bibliographystyle{ACM-Reference-Format}
\bibliography{refs}

\appendix

\section{Annotations} \label{annotation-process}

\subsection{Corpus selection} 
The selection criteria for the corpus are the same as in \citet{fabris2022algorithmic}. The overall scope of considered literature consists of all articles that were published in either (1) the proceedings of fairness-related conferences such as the ACM Conference on Fairness, Accountability, and Transparency (\textit{FAccT}) and the AAAI/ACM Conference on Artificial Intelligence, Ethics and Society (\textit{AIES}), (2) the proceedings of major machine learning conferences, including the IEEE/CVF Conference on Computer Vision and Pattern Recognition (\emph{CVPR}), the Conference on Neural Information Processing Systems (\emph{NeurIPS}), the International Conference on Machine Learning (\emph{ICML}), the International Conference on Learning Representations (\emph{ICLR}), the ACM SIGKDD International Conference on Knowledge Discovery and Data Mining (\emph{KDD}), or (3) the proceedings of any of the ``Past Network Events`` or ``Older Workshops'' as listed on the FAccT Network. Works from 2014 up to and including June 2021 were considered (including \textit{FAccT}, \textit{ICLR}, \textit{AIES} and \textit{CVPR} in 2021). This list of literature was narrowed down to fairness-related articles by a manual review, after first filtering for articles which included one of the following substrings in their titles (with * denoting wildcards): *fair*, *bias*, discriminat*, *equal*, *equit*, disparate, *parit*.

\subsection{Annotation Process}
Annotations were performed by the first author and two research assistants over the course of multiple months. Research assistants were fairly compensated for their work, following university guidelines at 12.00 EUR per hour without an academic degree and 14.00 EUR per hour with a Bachelor's degree. The annotation scheme and process were developed by the authors and research assistants received interactive training on the annotation process.

Annotations were made using Google Sheets using two tables: \textit{Datasets} and \textit{Datasets-x-Papers}, with each annotated column having an explanatory note regarding its annotation scheme. Datasets were randomly assigned to annotators, based on their internal unique identifiers. Dataset-x-Paper combinations were assigned based on assigned datasets, with a subset of them being reassigned based on annotator availability towards the end of the annotation process.

Throughout the annotation process there were weekly meetings to address any difficulties or ambiguities with annotations and the additional option for asynchronous discussion via chat software. Difficult annotations could be marked as requiring additional input. Additionally, annotation quality was checked on face validity by the first author for a subset of annotations.

\subsection{Annotation Instructions}

Besides in-person training on the annotation process, the following written instructions were made available to annotators:

\namedpar{Tables}

\begin{itemize}
    \item \emph{Datasets}, which contains data on individual datasets, incl. any varieties
    \item \emph{Datasets-x-Papers}, which contains an entry for every dataset and paper that makes use of said dataset.
\end{itemize}

\namedpar{Annotation process}.
Start by annotating the data for a dataset, then annotate the papers that use it. Update the entry of the dataset if changes become necessary.
For every column, you can find information on how to annotate it by hovering over its title.
Annotate each row from left to right.
When you want to put multiple values in a single cell (e.g. multiple column names), separate them with semicolons.
When any questions emerge or something is unclear, post in the slack channel.
Please always use filter views when performing annotations, to only see the annotations assigned to you.

\namedpar{Standardized Process for Searching relevant sections}.
When annotating entries in \emph{Datasets-x-Papers}, it's important we do our due diligence in searching for information about how a dataset was used.
This is especially important in regards to a paper's code (as code is typically an external ressource, so easier to miss). 
Please always try at least the following 5 steps when searching information about how a dataset is used.
You're also free to try additional ways of finding information about the dataset, but we want to make sure, that at least these steps have been performed for every paper.

\namedpar{Searching for Code}

\begin{enumerate}
    \item Search for "github" and "gitlab" in the paper.
    \item Search for the paper's name on google.
        Sometimes there's an external repository with code that uses the paper's name, but is not referenced in the paper.
    \item Check in the official location of the paper whether it has supplementary material e.g. an appendix or zip files. These can contain code or a detailed description of datasets.
\end{enumerate}
\namedpar{Finding relevant sections}
\begin{enumerate}
    \item Search for the common names of the dataset itself to find information about it (if it has a common name)
    \item Search for "dataset" or "data" to find the relevant sections describing how data is used.
\end{enumerate}

\section{Robustness}\label{app:robustness}
In this appendix, we investigate the robustness of Section \ref{sec:neglected} findings across time, fairness tasks, and beyond tabular datasets. Additionally, we ensure that the tabular datasets we focused on remained central in the literature. Considering the most recent proceedings (2023) of two well-known machine learning and fairness conferences such as \emph{ICML} and \emph{FAccT}, we select all articles whose titles contain the string \texttt{fair}. We manually select articles that focus on quantitative analyses of group fairness, without any restriction based on task or data specification. For each of these manuscripts, we annotate dataset and protected attribute usage. Our findings are presented below.

\namedpar{Popular datasets remained popular}. Our analysis in Section \ref{sec:neglected} is based on publications up to 2021, building on top of \citet{fabris2022algorithmic}. We find that 8 out of 10 most popular datasets remain the same, with the key exception of the recently-introduced Folktables datasets \citep{ding2021retiring} (10 usages), complementing but not \emph{retiring} Adult (13 usages). All such datasets are tabular, confirming the centrality of this data modality in fair ML research.

\begin{table}
\caption{\textbf{The usage of datasets remained highly similar in 2023.} Usage of datasets
in fairness-related articles published at FAccT and ICML 2023 compared
to usage within the annotated literature. Only datasets which are used
at least twice in 2023 are shown. Datasets are ordered by their usage in
2023.}\tabularnewline

\centering
\begin{tabular}[t]{llll|lll}
\hline
\multicolumn{1}{c}{ } & \multicolumn{3}{c}{2023} & \multicolumn{3}{c}{Up to 2021} \\
\cline{2-4} \cline{5-7}
Dataset Name & Rank & Fraction & N & Rank & Fraction & N \\
\hline
Adult & 1 & 20.3\% & 13 & 1 & 30.0\% & 84\\
Folktables \textit{(new dataset)} & 2 & 15.6\% & 10 & - & - & -\\
COMPAS & 3 & 12.5\% & 8 & 2 & 24.6\% & 69\\
Communities; Communities and Crime & 4 & 7.8\% & 5 & 4 & 4.3\% & 12\\
German; German Credit; Credit & 5 & 6.2\% & 4 & 3 & 9.3\% & 26\\
Law\_School & 5 & 6.2\% & 4 & 4 & 4.3\% & 12\\
Bank; Bank Marketing; Marketing & 7 & 4.7\% & 3 & 6 & 3.2\% & 9\\
default of credit card clients & 8 & 3.1\% & 2 & 11 & 1.4\% & 4\\
Student; Student Performance & 8 & 3.1\% & 2 & 21 & 0.4\% & 1\\
\hline
\end{tabular}
\end{table}

\begin{figure}%
    \includegraphics[width=0.75\textwidth]{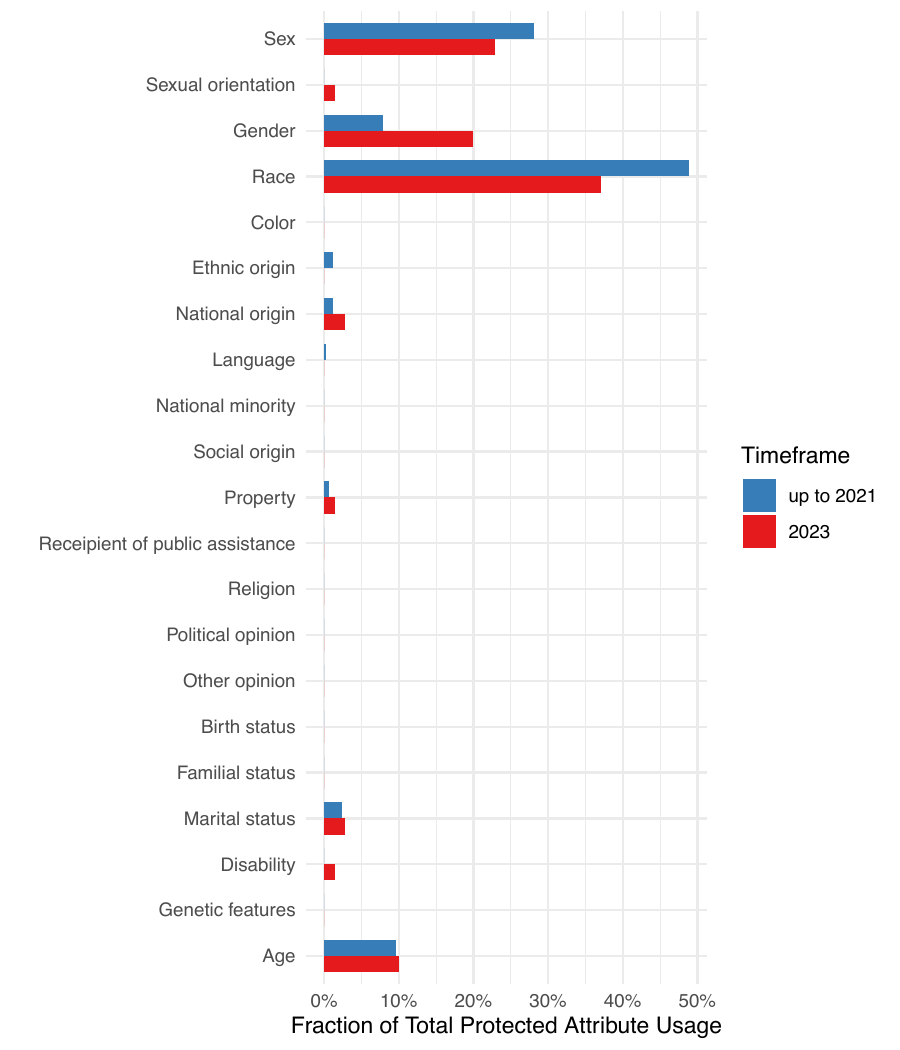} 
    \caption{\textbf{The usage of protected attributes remained similar in 2023.} Relative usage of protected attributes in the annotated literature up to 2021 and within the subset of literature we examined in 2023. Usage within the annotated literature corresponds to the right half of Figure \ref{fig:sensitive-attributes}.}%
    \label{fig:bar-sensitive-2023}%
\end{figure}

\namedpar{Neglected identities remain neglected}. 
Figure \ref{fig:bar-sensitive-2023} compares protected attributes in fair ML experiments up to 2021 and in 2023. Although we find isolated experiments on sexual orientation, property, and disability, it is clear that these attributes, as well as religion ($n=0$), remain understudied, especially in comparison with sex, gender, and race.  It is worth noting that we follow the naming of manuscript authors and dataset creators for sex and gender; the drop of the former in favor of the latter is a consequence of this fact and may not reflect an actual focus shift.

\newpage

\section{Addendum: Opaque Preprocessing of Bank} \label{app:opaque}

Here we present supplementary figures and information for the analyses in Section~\ref{sec:opaque}. The performance metrics used in this work are accuracy (Eq \ref{eq:acc}), balanced accuracy (Eq \ref{eq:bacc}), and F1 score (Eq \ref{eq:f1}).

\begin{align}
    \text{Precision} &= \Pr(y=1|\hat{y}=1) \nonumber \\
    \text{Recall} &= \Pr(\hat{y}=1|y=1) \nonumber \\
    \text{Specificity} &= \Pr(\hat{y}=0|y=0) \nonumber \\
    \text{Acc} &= \Pr(\hat{y}=y)  \label{eq:acc} \\
    \text{bACC} &= \frac{\text{Specificity} + \text{Recall}}{2}  \label{eq:bacc} \\
    \text{F1 Score} &= \frac{2}{\text{Precision}^{-1}+\text{Recall}^{-1}} \label{eq:f1}
\end{align}

The fairness metrics used in this work are equalized odds difference  (Eq \ref{eq:eod}), demographic parity difference (Eq \ref{eq:dpd}), and disparate impact (Eq \ref{eq:di}).

\begin{align}
    \text{EOD} &= \max_g \Pr(\hat{y}=1|y=1, S=g) - \min_g \Pr(\hat{y}=1|y=1, S=g)\label{eq:eod} \\
    \text{DPD} &= \max_g \Pr(\hat{y}=1|S=g) - \min_g \Pr(\hat{y}=1|S=g)\label{eq:dpd} \\
    \text{DI} &= \frac{\max_g \Pr(\hat{y}=1|S=g)}{ \min_g \Pr(\hat{y}=1|S=g)} \label{eq:di} 
\end{align}

\begin{figure}%
    \includegraphics[width=0.95\textwidth]{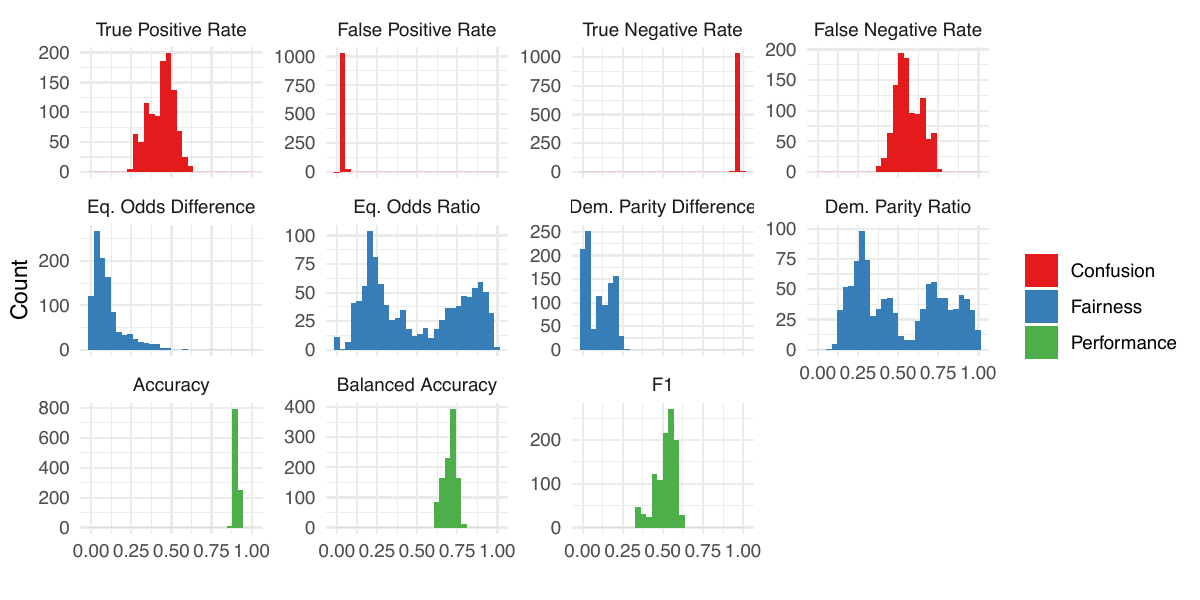} 
    \caption{\textbf{There is a large degree of overall variation, especially on fairness metrics.} Histograms displaying the overall variation on different metrics within and across different scenarios and repetitions of the analysis.}%
    \label{fig:fig-hist-choices-variation}%
\end{figure}

The overall variation of different metrics for the first experiment in Section~\ref{sec:opaque} is illustrated in Figure~\ref{fig:fig-hist-choices-variation}. As can be seen, there exists ample variation across the different metrics and variation is especially pronounced on metrics of algorithmic fairness.

\begin{figure}%
    \includegraphics[width=0.7\textwidth]{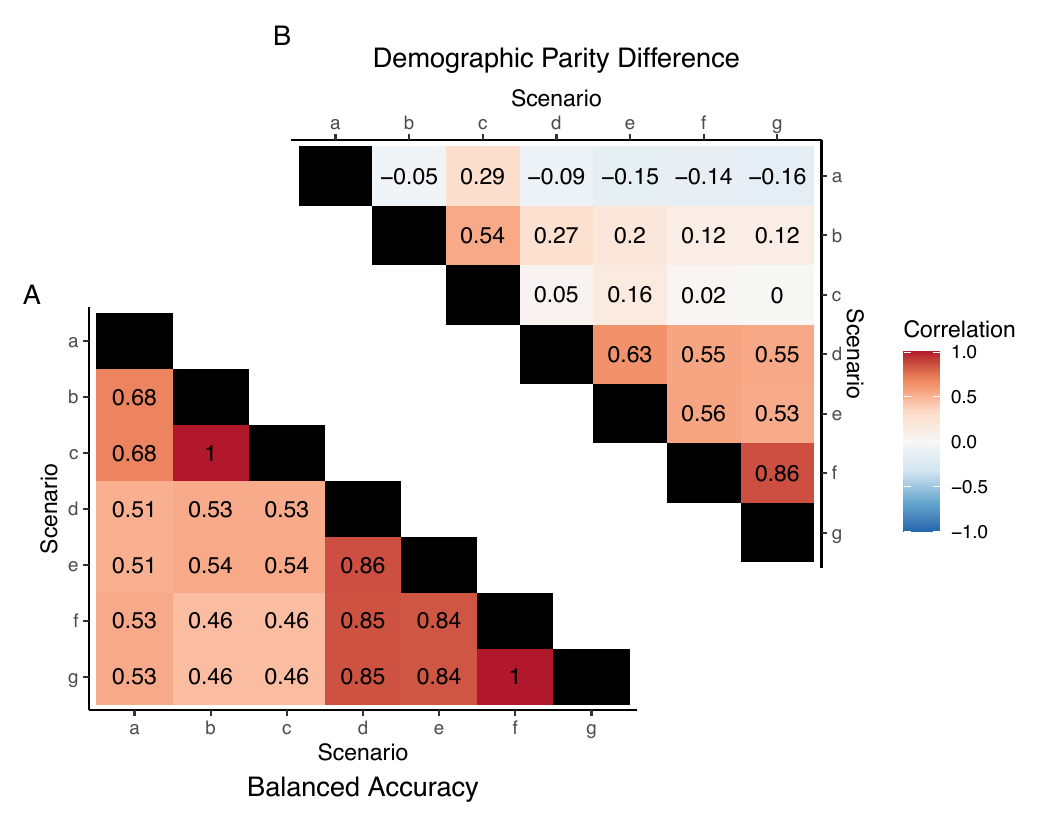} 
    \caption{Spearman’s $\rho$ correlations of model ranks on (A) Balanced Accuracy and (B) Demographic Parity Difference between different scenarios. Letters correspond to the scenarios described in Figure~\ref{fig:sankey}.}%
    \label{fig:corplot-balacc-demparity}%
\end{figure}

Figure \ref{fig:corplot-balacc-demparity} depicts correlation matrices for the first experiment in Section \ref{sec:opaque}, with different performance and fairness measures, namely \emph{balanced accuracy} and \emph{demographic parity difference}. Although we still note instability in fairness-based model comparison, comparisons based on demographic parity are more stable than for equalized odds difference. We interpret this as a consequence of a classifier's (groupwise) acceptance rate $\Pr(\hat{y}=1)$ being more stable than its (groupwise) true positive rate $\Pr(\hat{y}=1|y=1)$ since the former is computed over all points in the test set, while the latter only on the positives ($y=1$). 

\begin{figure}%
    \includegraphics[width=0.7\textwidth]{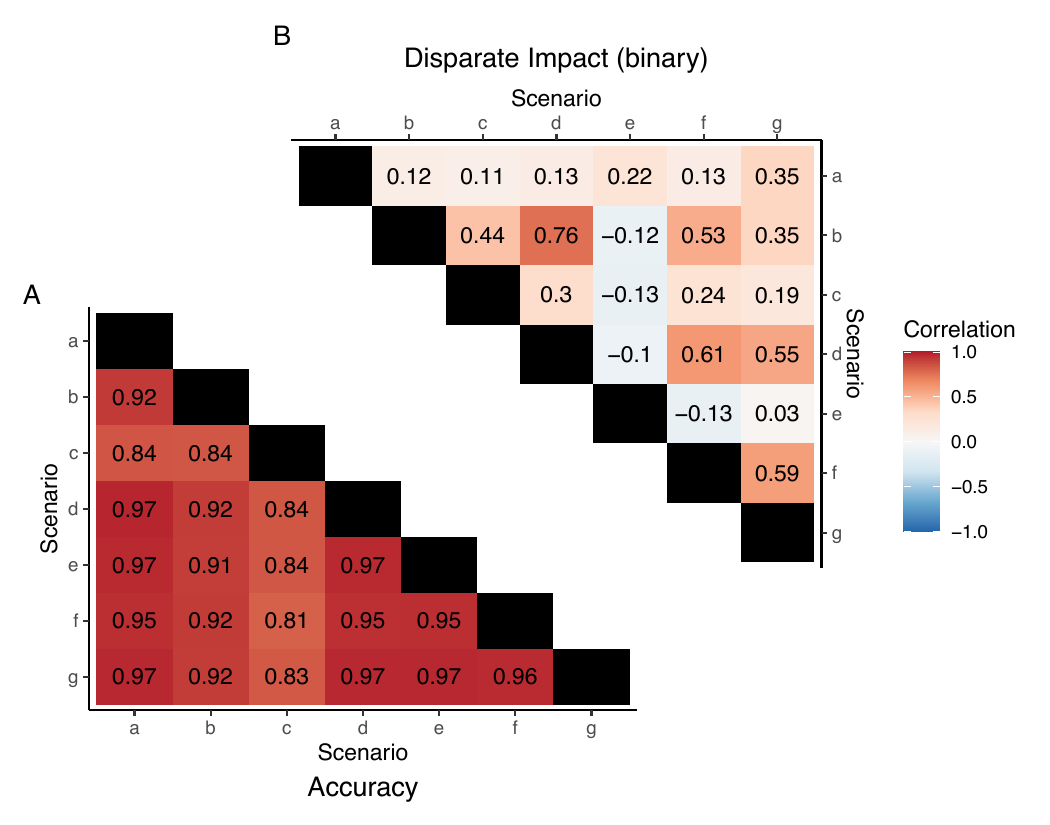} 
    \caption{Spearman’s $\rho$ correlations of model ranks on (A) Raw Accuracy and (B) Disparate Impact (binary) between different scenarios when reproducing our analysis from Section~\ref{sec:opaque} using an existing selection of fairness-aware algorithms and methodology \citep{friedler2019comparative}. Letters correspond to the scenarios described in Figure~\ref{fig:sankey}.}%
    \label{fig:corplot-friedler}%
\end{figure}

For the second experiment in the section, we aimed to repeat our analysis replicating a highly popular setting. We therefore used the same selection of (mainly) fairness-aware algorithms used in \citet{friedler2019comparative} and applied their methodology on the differently processed versions of the Bank dataset in Figure~\ref{fig:sankey}. Specifically, we used the \textit{numeric} variant of their analysis, as it works with a sufficiently large selection of algorithms and does not require the protected attribute to be binary. The correlation matrices for \emph{accuracy} and \emph{disparate impact} across scenarios are depicted in Figure~\ref{fig:corplot-friedler}. Both metrics were chosen following \citet{friedler2019comparative}. Disparate impact is calculated using a binary version of the protected attribute, split into privileged and unprivileged groups. Using the non-binary, averaged version of disparate impact also discussed in the original paper, lead to similar and even more diverse results.

\end{document}